\documentclass[letterpaper, 10 pt, conference]{ieeeconf}  

\IEEEoverridecommandlockouts                              

\usepackage{url}
\usepackage{cite}
\usepackage{balance}
\usepackage{graphicx}
\usepackage{amsfonts}
\usepackage{fancyhdr}
\usepackage{comment}
\usepackage{times}
\usepackage{amsmath}
\usepackage{changepage}
\usepackage{amssymb}
\usepackage{enumerate} 
\usepackage{algorithmic}
\usepackage{bm}
\usepackage{calligra}
\usepackage{multirow}
\usepackage{tabularx}

\usepackage{booktabs}
\usepackage{mathtools}
\usepackage{arydshln}
\usepackage{latexsym} 
\usepackage{amsmath}
\usepackage{amssymb}
\usepackage{color}
\usepackage[ruled,vlined]{algorithm2e}
\usepackage{float}
\usepackage{tabstackengine}
\usepackage{siunitx}
\usepackage{colortbl}
\usepackage{hhline}
\usepackage{bookmark}
\usepackage{dirtree}
\usepackage{subfigure}

\newcommand{\rt}{\textcolor[rgb]{1,0,0}}

\begin{document}
\title{\LARGE \bf 
TAIL: A Terrain-Aware Multi-Modal SLAM Dataset for Robot Locomotion in Deformable Granular Environments
}

\author{
Chen Yao$^{*}$$^{1}$, Yangtao Ge$^{*}$$^{1}$, Guowei Shi$^{*}$$^{1}$,  Zirui Wang$^{*}$$^{1}$, Ningbo Yang$^{1}$, Zheng Zhu$^{1}$, \\ Hexiang Wei$^{2}$, Yuntian Zhao$^{1}$, Jing Wu$^{1}$, Zhenzhong Jia$^{1**}$
\thanks{$^{1}$Shenzhen Key Laboratory of Biomimetic Robotics and Intelligent Systems, Department of Mechanical and Energy Engineering, Southern University of Science and Technology (SUSTech), Shenzhen, 518055, China. Guangdong Provincial Key Laboratory of Human-Augmentation and Rehabilitation Robotics in Universities, SUSTech, Shenzhen, 518055, China. }
\thanks{$^{2}$The Hong Kong University of Science and Technology (Guangzhou), Nansha, Guangzhou, 511400, Guangdong, China.}
\thanks{This work was supported in part by the Guangdong Natural Science Fund-General Programme under grant no. 2021A1515012384, the Science, Technology and Innovation Commission of Shenzhen Municipality under grant no. ZDSYS20200811143601004, and the National Science Foundation of China(NSFC)U1913603. }
\thanks{$^{*}$Equal contribution.}
\thanks{$^{**}$Corresponding author: {\tt\small jiazz@sustech.edu.cn}}
}

\maketitle

\begin{abstract}
Terrain-aware perception holds the potential to improve the robustness and accuracy of autonomous robot navigation in the wilds, thereby facilitating effective off-road traversals.
However, the lack of multi-modal perception across various motion patterns hinders the solutions of Simultaneous Localization And Mapping (SLAM), especially when confronting non-geometric hazards in demanding landscapes.
In this paper, we first propose a Terrain-Aware multI-modaL (TAIL) dataset tailored to deformable and sandy terrains. 
It incorporates various types of robotic proprioception and distinct ground interactions for the unique challenges and benchmark of multi-sensor fusion SLAM. 
The versatile sensor suite comprises stereo frame cameras, multiple ground-pointing RGB-D cameras, a rotating 3D LiDAR, an IMU, and an RTK device. This ensemble is hardware-synchronized, well-calibrated, and self-contained.
Utilizing both wheeled and quadrupedal locomotion, we efficiently collect comprehensive sequences to capture rich unstructured scenarios. 
It spans the spectrum of scope, terrain interactions, scene changes, ground-level properties, and dynamic robot characteristics.
We benchmark several state-of-the-art SLAM methods against ground truth and provide performance validations. Corresponding challenges and limitations are also reported.
All associated resources are accessible upon request at \url{https://tailrobot.github.io/}. 
\end{abstract}
\vspace{-3mm}

\section{Introduction}
\label{sec:introduction}

Autonomous robotic platforms have been widely applied in outdoor missions such as exploration due to their promising mobility and reliability.
However, these scientific-rich regions are often covered with rugged off-road terrains, which could lead to traversal hazards like sinkage or slippage, especially in deformable, granular scenes \cite{webster2009nasa, kerr2009mars}.
Recognizing this, terrain-aware technologies are gaining popularity, empowering various robots to explore unexpected fields and providing reliable solutions for traversing uncertain risks.
Multi-sensor fusion holds strong potential in enabling safe terrain traversal for field robotic applications. Integrating multi-modal data would ensure accurate and robust perception on deformable, granular terrains, allowing for the understanding of dynamic robot-ground interaction and enhancement of traversability \cite{fan2021step}. 
Additionally, these sensors often vary in characteristics and can guarantee system redundancy in a complementary way.
Recent research on critical multi-sensor SLAM has progressively emerged as a necessary tool for negotiating such challenging terrains. 

\begin{figure}[t!]
\vspace{0.2cm}
\centering
\includegraphics[scale=0.34]{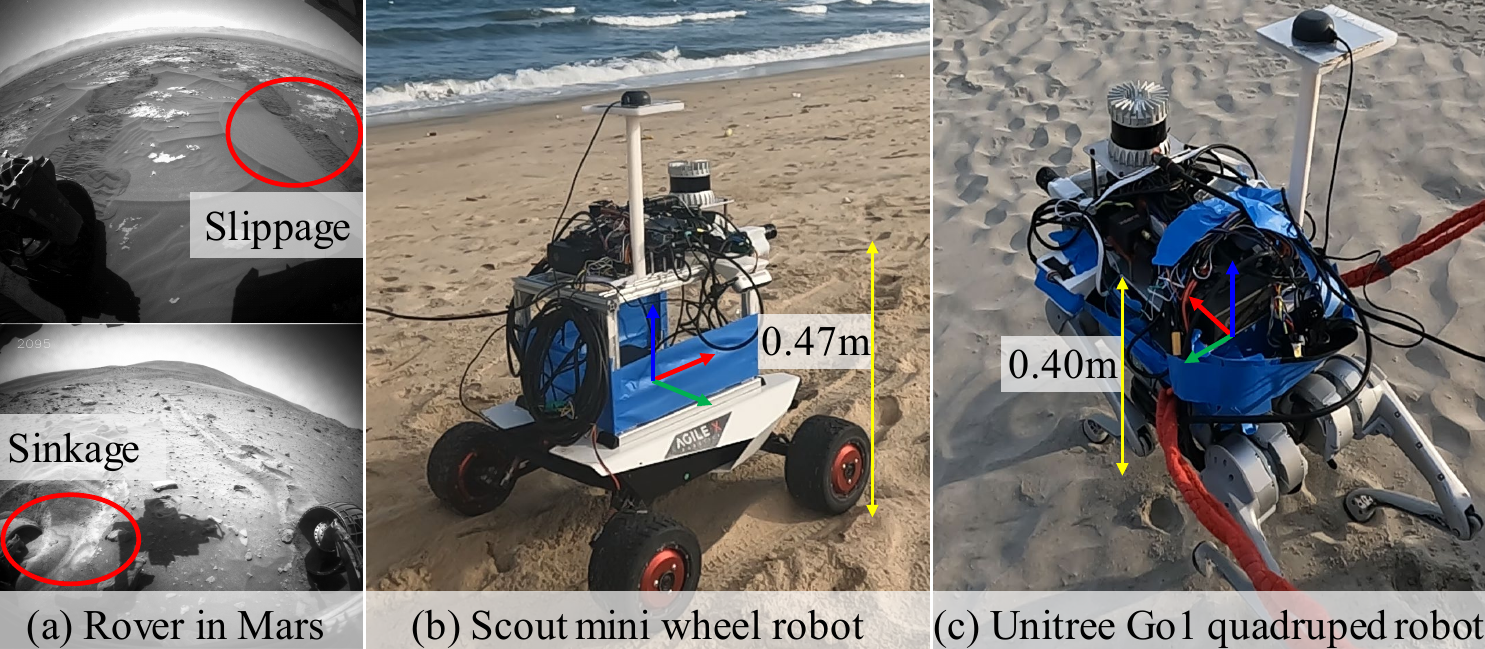}

\vspace{-0.2cm}
\caption{ (a) Terrain hazards encountered by NASA \emph{Spirit} rover \cite{smith2009new}. To study terrain-related hazards, we build a versatile sensor suite to record different data sequences in challenging soft, granular terrains. 
Both (b) wheeled robot and (c) quadruped robot platforms are used for carrying the designed sensor suite during terrain traversals of diverse sandy terrains, where robot locomotion can become quite challenging in certain scenarios.
}
\label{fig:intro_picture}
\vspace{-0.5cm}
\end{figure}

In recent years, the growing demand for effective SLAM solutions \cite{shan2021lvi} has prompted the development of multi-sensor datasets.  
These datasets offer fair and repeatable operational tools, serving as valuable resources for algorithm implementation, comparison, and validation.
Additionally, they can effectively evaluate challenging scenarios to analyze the diverse robot characteristics and limitations of current SLAM techniques.
Furthermore, they could help to reduce unnecessary financial and engineering requirements.
Certain datasets also introduce novel sensors, challenging existing solutions and fostering further research potentials  \cite{mueggler2017event}.
Consequently, the use of such datasets accelerates breakthroughs concerning accuracy, robustness, and computational performance, thereby validating and enhancing general robot autonomy.
 
\begin{table*}[t!]
\vspace{0.3cm}
\centering
\caption{Comparison with various notable public datasets.}
\renewcommand\arraystretch{1.0}
\renewcommand\tabcolsep{2.0pt}
\scriptsize
\begin{tabular}{lllcccccccc}
\toprule[0.03cm]
 &  &  & \multicolumn{6}{c}{Sensors} &  &  \\ \cline{4-9}
\multirow{-2}{*}{Dataset} & \multirow{-2}{*}{Platform} & \multirow{-2}{*}{Environments} & IMU & LiDAR & GPS & Frame Cam. &  Down'd RGBD Cam. & Robot Status & \multirow{-2}{*}{Sync.} & \multirow{-2}{*}{GT Pose} \\ \toprule[0.03cm]
\cellcolor[HTML]{D8D8D8}EuRoc \cite{burri2016euroc} & MAV & Indoors & \checkmark &  &  & \checkmark &  &  & Hw & Mocap/LT \\
\cellcolor[HTML]{D8D8D8}KAIST \cite{jeong2018complex} & Car & Outdoors-Urban & \checkmark & \checkmark &  &  &  &  & \textbackslash{} & SLAM \\
\cellcolor[HTML]{D8D8D8}KITTI \cite{geiger2013vision} & Car & Outdoors-Urban & \checkmark & \checkmark & \checkmark & \checkmark &  &  & Hw/Sw & RTK-GPS \\
\cellcolor[HTML]{D8D8D8}M2DGR \cite{yin2021m2dgr} & UGV & Campus/Lab & \checkmark & \checkmark & \checkmark & \checkmark &  &  & Sw & RTK-GPS/LT/Mocap \\
\cellcolor[HTML]{D8D8D8}FusionPortable \cite{jiao2022fusionportable} & Handheld/QR/UGV & In/Outdoors & \checkmark & \checkmark & \checkmark & \checkmark &  &  & Hw & Mocap/RTK-GPS/SLAM \\
\cellcolor[HTML]{D8D8D8}VECTOR \cite{gao2022vector} & Handheld/hemelet & Indoors & \checkmark & \checkmark &  & \checkmark &  &  & Hw & MoCap + SLAM \\
SubT-MRS \cite{zhao2023subtmrs} & Handheld/QR/UGV/MAV & In/Outdoors & \checkmark & \checkmark &  & \checkmark &  &  & Hw & Super Odometry \cite{zhao2021super} \\
Wild-Places \cite{knights2023wild} & Handheld & Outdoors-Jungle & \checkmark & \checkmark &  & \checkmark &  &  & Hw & Wildcat SLAM \cite{ramezani2022wildcat} \\
DiTer \cite{jeongditer}  & QR & Outdoors & \checkmark & \checkmark & \checkmark & \checkmark & \checkmark & Partial & \textbackslash{} & RTK-GPS \\
MADMAX \cite{meyer2021madmax} & UGV & Outdoors-Desert & \checkmark &  & \checkmark & \checkmark &  &  & Hw/Sw & RTK-GPS \\
Katwijk \cite{hewitt2018katwijk} & Rover & Outdoors-Beach & \checkmark & \checkmark & \checkmark & \checkmark & \checkmark &  & \textbackslash{} & RTK-GPS \\ \midrule[0.03cm]
\textbf{Ours (TAIL)} & UGV/QR & Soft terrains & \checkmark & \checkmark & \checkmark & \checkmark & \checkmark & \checkmark & Hw & RTK-GNSS \\ \bottomrule[0.03cm]	
\multicolumn{11}{l}{
RTK: Real-time kinematic.
QR: Quadruped robot.
Mocap: Motion capture system. 
LT: Laser tracker.
Sw/Hw: software/hardware synchronization.
Down’d: Downward.
}	
\end{tabular}
\label{tab:related_work_dataset}
\vspace{-0.5cm}
\end{table*}

\vspace{-2mm}
\subsection{Related Work}
\label{sec:related_work} 
\vspace{-1mm}

Over the past years, numerous SLAM datasets, employing both simple and multi-sensor configurations, have been actively developed, covering structured and unstructured scenarios with unique robots, such as terrain hazards \cite{rogers2020test}.

Several datasets were collected using single or simple sensor combinations, such as inertial measurement units (IMU), global positioning systems (GPS), cameras, and light detection and ranging (LiDAR).  
Notably, LiDAR-inertial odometry (LIO) and Visual-inertial odometry (VIO) are designed to fuse cameras or LiDAR data with inertial measurements to improve the robustness and accuracy of localization.
For example, EuRoc dataset \cite{burri2016euroc} tracked a drone in both room and industrial environments.
KAIST \cite{jeong2018complex} presented an urban car dataset to capture the genuine features in areas where GPS signals show apparent degradation.
However, these limited data could not capture full challenges in practice.

Several multi-sensor datasets are obtained in structured or semi-structured environments to address challenges.
KITTI dataset \cite{geiger2013vision} presented extensive car driving tasks to explore different perception scenes.
M2DGR dataset \cite{yin2021m2dgr} employed a ground robot to record different challenging scenarios.
FusionPortable \cite{jiao2022fusionportable} proposed a campus-scene dataset tailored for evaluating SLAM across multi-platforms.
VECtor dataset \cite{gao2022vector} covered indoor scenarios and served as a versatile benchmark for event-centric studies with different devices.

Moreover, multi-sensor datasets prove essential for implementation in unstructured fields characterized by complex and natural obstacles. 
SubT-MRS \cite{zhao2023subtmrs} showed their real-world collection of datasets obtained from Subterranean (SubT) environments, with a focusing theme on testing SLAM in multi-robot, multi-spectral-inertial, and multi-degraded scenarios.
Wild-Places \cite{knights2023wild} introduced a large-scale dataset in natural unstructured environments captured through a handheld device.
DiTer \cite{jeongditer} presented a terrain dataset designed for ground mapping using a quadruped robot.
MADMAX dataset \cite{meyer2021madmax} utilized a UGV in the Moroccan desert to record different visual-inertial sequences.
Katwijk \cite{hewitt2018katwijk} described a multi-sensor dataset along the beach near Katwijk to emulate potential rover landing sites.

This paper focuses on multi-modal SLAM datasets in unstructured granular terrain environments (see Fig.\ref{fig:intro_picture}), where geometric features are sparse and terrain-aware technology plays a pivotal role. 
Tab.~\ref{tab:related_work_dataset} reviews the representative datasets related to our work.
We see that datasets focusing on deformable terrain traversals (MADMAX and Katwijk) only studied wheeled locomotion, while those considered legged robots (FusionPortable, SubT-MRS, and DiTer) did not investigate soft environments. 
These works consider unique platforms in real-world operations, which may not always yield highly feasible measurements. However, the provision of intrinsic profiles can indeed enhance dynamic interactions and handle locomotion challenges in a complementary way\cite{yao2023adaptive}.
These problems motivate us to develop a dataset for distinct wheeled and legged locomotion in soft terrains and benchmark SLAM solutions with multiple sensory configurations.
We expect that our dataset will provide significant advancements to bridge the gap in multi-sensor fusion within the SLAM and field robotics research community.

\vspace{-2mm}
\subsection{Contributions}
\vspace{-1mm}

We create the \textbf{T}errain-\textbf{A}ware Mult\textbf{I-}Moda\textbf{L} \textbf{(TAIL)} dataset, a viable and promising SLAM dataset for both wheeled and legged robots traveling dynamically in deformable, granular scenarios, as described in Fig. \ref{fig:intro_picture}.
The main purpose of TAIL is to propel the development of multi-sensor fusion SLAM techniques in soft terrains, which can capture the distinct heterogeneous proprioception and exteroception.
Specially, it exhibits various kinematic patterns and multiple sandy terrestrial interaction to challenge SLAM algorithms.

Our main contributions are listed as follows:
\begin{itemize}
\item We construct the TAIL dataset, which packages a versatile and hardware-synchronized sensor suite capable of millisecond-level and adjustable time intervals, comprising comprehensive sensors, a processor, and a battery (Fig. \ref{fig:sensor_picture}). This device can serve as an experimental tool for algorithm evaluation on sandy terrains.

\item We collect the TAIL sequences with rich info. covering a wide spectrum of field complexities (texture-less), scope (surrounding and ground-pointing), and scene changes (illumination, moving objects, and flowing sands) faced in two types of soft sands.
Notably, we are the first to present distinct proprioception for wheeled and quadruped robots that can capture distinct motion interaction within two changeable sandy scenes.

\item We conduct algorithm validation using the TAIL dataset to benchmark representative state-of-the-art (SOTA) SLAM algorithms. 
The results indicate that multi-terrains and distinct robot locomotion modes pose additional challenges to SLAM, emphasizing the need for heightened attention in designing sensor fusion SLAM.

\end{itemize}

The rest of this paper is organized as follows:
Sec.~\ref{sec:system} introduces the sensor suite system.
The detailed features of the dataset sequences are outlined in Sec.~\ref{sec:dataset}.
In Sec.~\ref{sec.experiment}, we benchmark the performance of various SLAM algorithms.
Conclusion and future work are discussed in Sec.~\ref{sec.conclusion}.

\begin{table*}[t!]
\vspace{0.3cm}
\centering 
\caption{Overview of the hardware specifications used in the multi-sensor device}
\setlength{\tabcolsep}{2.0mm}{
\resizebox{2.0\columnwidth}{!}{
\begin{tabular}{ccclll}
\toprule
Hardware & Type & Characteristics & Rate & Topic name & Topic type \\ \midrule[0.03cm]
\multirow{6}{*}{\begin{tabular}[c]{@{}c@{}}3D \\ LiDAR\end{tabular}} & \multirow{6}{*}{Ouster OS0-128} & \multirow{6}{*}{\begin{tabular}[c]{@{}c@{}}128×2048 points, 50m range.\\ ±1.5-5cm range precision.\\ FOV: 90$^{\circ}$V/360$^{\circ}$H.\\ IMU: ICM20948,100Hz, 9-axis MEMS.\end{tabular}} & \multirow{6}{*}{10Hz} & /ouster/points & sensor\_msgs/PointCloud2 \\
 &  &  &  & /ouster/nearir\_image & sensor\_msgs/Image \\
 &  &  &  & /ouster/range\_image & sensor\_msgs/Image \\
 &  &  &  & /ouster/reflec\_image & sensor\_msgs/Image \\
 &  &  &  & /ouster/signal\_image & sensor\_msgs/Image \\
 &  &  &  & /ouster/imu & sensor\_msgs/Imu \\ \midrule[0.01cm]
\multirow{2}{*}{\begin{tabular}[c]{@{}c@{}}Frame\\  Cameras\end{tabular}} & \multirow{2}{*}{\begin{tabular}[c]{@{}c@{}}2$\times$ HKVision \\ MV-CA013-A0UC\end{tabular}} & \multirow{2}{*}{\begin{tabular}[c]{@{}c@{}}1280×1024 pixels,global shutter.\\ FOV: 66:5$^{\circ}$vert., 82:9$^{\circ}$horiz, \textgreater{}53dB.\end{tabular}} & \multirow{2}{*}{30Hz} & /left\_camera/image/compressed & sensor\_msgs/CompressedImage \\
 &  &  &  & /right\_camera/image/compressed & sensor\_msgs/CompressedImage \\ \midrule[0.01cm]
\multirow{5}{*}{\begin{tabular}[c]{@{}c@{}}RGB-D\\  Camera\end{tabular}} & \multirow{5}{*}{Azure Kinect DK} & \multirow{5}{*}{\begin{tabular}[c]{@{}c@{}}1280×720 pixels, time-of-flight.\\ FOV: 65$^{\circ}$V/75$^{\circ}$H. 0.5-3.86m. \\ Depth: 640×576 pixels \\ Instrinsic calibrated.\\ Ground-level view. \\ \end{tabular}} & \multirow{5}{*}{30Hz} & /depth/camera\_info & sensor\_msgs/CameraInfo \\
 &  &  &  & /depth/image\_raw & sensor\_msgs/Image \\
 &  &  &  & /depth/points & sensor\_msgs/PointCloud2 \\
 &  &  &  & /rgb/camera\_info & sensor\_msgs/CameraInfo \\
 &  &  &  & /rgb/image\_raw/compressed & sensor\_msgs/CompressedImage \\ \midrule[0.01cm]
\multirow{8}{*}{\begin{tabular}[c]{@{}c@{}}RGB-D\\  Camera\end{tabular}} & \multirow{8}{*}{\begin{tabular}[c]{@{}c@{}}2$\times$ Intel \\ Realsense D435i\end{tabular}} & \multirow{8}{*}{\begin{tabular}[c]{@{}c@{}}640×480 pixels, HD 16:9. 0.3-3.0m. \\ FOV: 58$^{\circ}$V/85.2$^{\circ}$H.\\ Only depth camera is synchronized. \\ Depth: 848×480 pixels \\ Instrinsic calibrated. \\Ground-level view. \\\end{tabular}} & \multirow{8}{*}{15Hz} & /D435\_left/color/camera\_info & sensor\_msgs/CameraInfo \\
 &  &  &  & /D435\_left/color/image\_raw/compressed & sensor\_msgs/CompressedImage \\
 &  &  &  & /D435\_left/depth/camera\_info & sensor\_msgs/CameraInfo \\
 &  &  &  & /D435\_left/depth/image\_rect\_raw & sensor\_msgs/Image \\
 &  &  &  & /D435\_right/color/camera\_info & sensor\_msgs/CameraInfo \\
 &  &  &  & /D435\_right/color/image\_raw/compressed & sensor\_msgs/CompressedImage \\
 &  &  &  & /D435\_right/depth/camera\_info & sensor\_msgs/CameraInfo \\
 &  &  &  & /D435\_right/depth/image\_rect\_raw & sensor\_msgs/Image \\ \midrule[0.01cm]
\multirow{2}{*}{IMU} & \multirow{2}{*}{XSens MTi-680G} & \multirow{2}{*}{\begin{tabular}[c]{@{}c@{}}Roll/Pitch: 0.2° RMS, Yaw: 0.5° RMS.\\ Pos: 1cm+1ppm CEP, Vel: 0.05m/s RMS.\end{tabular}} & \multirow{2}{*}{100Hz} & \multirow{2}{*}{/imu/data} & \multirow{2}{*}{sensor\_msgs/Imu} \\
 &  &  &  &  &  \\ \midrule[0.01cm]
\multirow{2}{*}{RTK} & \multirow{2}{*}{Ublox ZED-F9P} & \multirow{2}{*}{\begin{tabular}[c]{@{}c@{}}6 DoF ground truth. INS: 9-axis \\ Localization accuracy 2cm.\end{tabular}} & 100Hz & /filter/positionlla & geometry\_msgs/Vector3Stamped \\ 
 &  &  & 1Hz & /status & diagnostic\_msgs/KeyValue \\ \midrule[0.01cm]
\multirow{4}{*}{\begin{tabular}[c]{@{}c@{}}Robot \\ platforms\end{tabular}} & \multirow{3}{*}{Scout Mini} & \multirow{3}{*}{\begin{tabular}[c]{@{}c@{}}4WD, 6 DOF wheel odometry.\\ Kinematic profiles, motor, driving status\end{tabular}} & \multirow{3}{*}{50Hz} & /cmd\_vel & geometry\_msgs/Twist \\
 &  &  &  & /odom & nav\_msgs/Odometry \\
 &  &  &  & /scout\_status & scout\_msgs/ScoutStatus \\ \cline{2-6} 
 & Unitree-GO 1 & \begin{tabular}[c]{@{}c@{}}6 DOF leg odometry.\\  Kinematic profiles, motor, gait, foot status\end{tabular} & 500Hz & /high\_state & unitree\_legged\_msgs/HighState \\ \bottomrule[0.03cm]
\end{tabular}}}
\label{tab:sensor_all}
\vspace{-0.5cm}
\end{table*}

\section{System Overview}
\label{sec:system}

\begin{figure}[t!]
\vspace{0.2cm}
\centering
\includegraphics[scale=0.32]{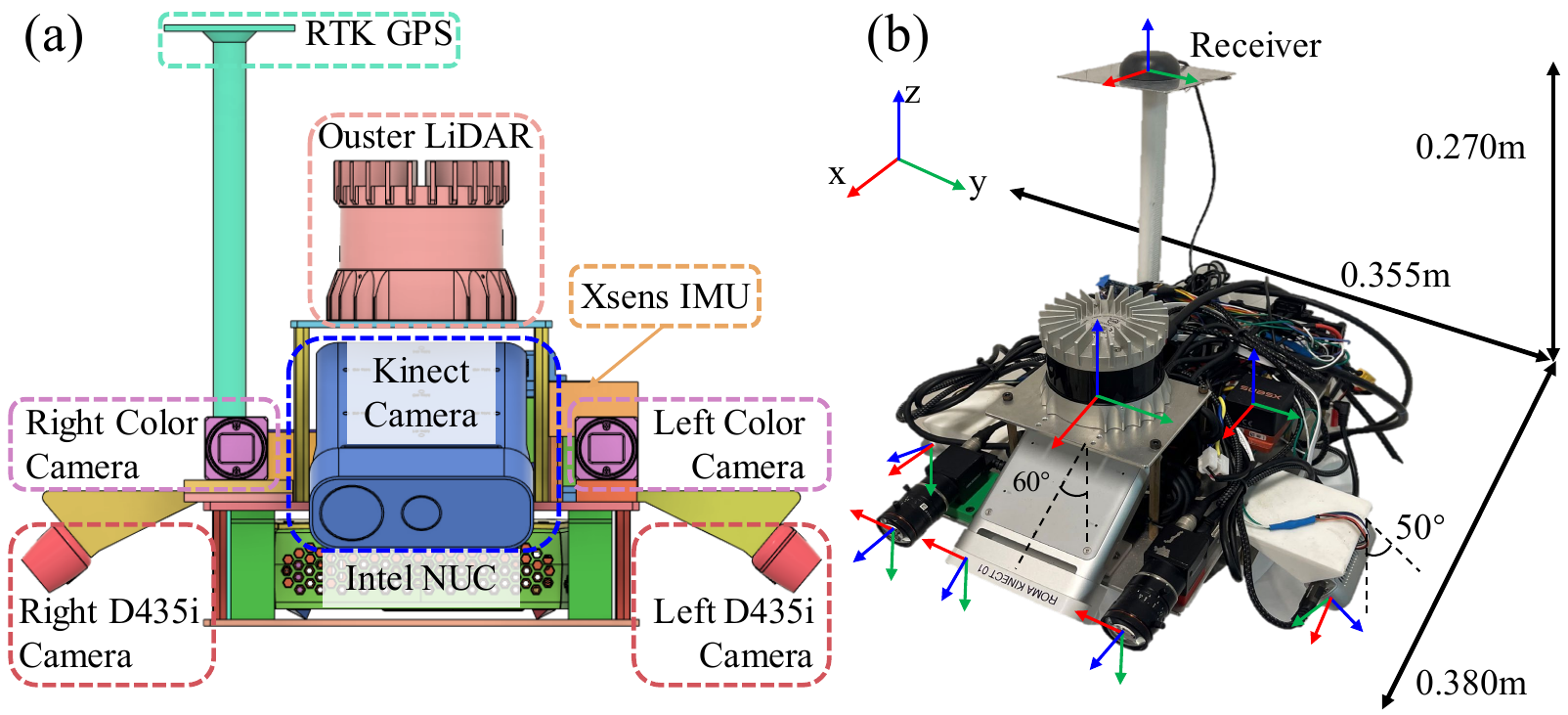}

\vspace{-2mm}
\caption{The specification for the multi-modal sensor suite. (a) CAD model and hardware components of the setup. (b) The mounted device's size and relevant coordinate frames of the sensor system.
High-resolution 3D LiDAR is located at the center and an RGBD camera is installed below. Two frames and RGBD cameras are mounted on the left and right sides, complemented by an internally placed high-precision IMU and a rear-installed RTK-GPS.
All these sensors are mounted on the same rigid aluminum frame and can be mounted on various robot platforms, e.g., see Fig. \ref{fig:intro_picture}. 
Thus, their spatial relations have no deviation, making it an effective test tool.
}
\label{fig:sensor_picture}
\vspace{-0.2cm}
\end{figure}

This section introduces the sensor suite configuration and provides detailed information on its hardware time synchronization and spatial-temporal calibration.
Fig. \ref{fig:sensor_picture} depicts the suite setup, with different sensors securely fastened to the supporting frame that can be mounted on different robots.

\subsection{Sensor Setup}
\label{sec:system_sensor_configuration}
We summarize the characteristics with corresponding messages for these sensor components in Tab. \ref{tab:sensor_all}.
All sensor drivers are implemented on an Intel NUC running Ubuntu 20.04 LTS and the Robot Operation System (ROS) Notietc system.
It is equipped with an i9 processor, a GeForce GTX 2060 GPU, $64$GB DDR4 memory, and $2$TB SSD.

\subsubsection{Inertial Measurement Unit}
We rigidly place the 9-axis IMU (model: Xsens MTi-680G) at the rear as the main inertial sensor.
Configured to operate at a high rate of $100$Hz, it provides low-drift data, effectively mitigating unexpected motion measurements.

\subsubsection{3D LiDAR}
We place the OS$0$-$128$ LiDAR on top of the suite to measure the large 3D surrounding environments.
We use four standoffs to raise the LiDAR's height to prevent possible obstruction of the bottom laser beams. 
Operating at $10$Hz, the LiDAR can output 1028$\times$128 near-IR images, range images, reflected images, and signal images, which correlate with point clouds of the same scenes. 
Additionally, the LiDAR's internal IMU can also output the linear accelerations and angular velocities at a rate of $100$Hz.

\subsubsection{Regular Stereo Frame Cameras}
A pair of HKVision MV-CA013-A0UC global-shutter color cameras are mounted as a forward-looking stereo camera system.
The horizontal baseline is approximately $16.7$mm. 
Each camera is paired with an MVL-HF0824M-10MP lens, offering a $62.46^\circ$$\times$$44.05^\circ$ field of view (FOV).
We set their exposure time as fixed values to reduce relative latency.

\subsubsection{RGB-D Cameras}
The Kinect Azure DK camera is installed in the middle of the suite, obtaining sufficient ground views by employing the unbinned narrow-FOV depth mode.
In addition, two Intel RealSense D435i RGB-D cameras are arranged on the side to enhance terrestrial observations. 
Note that only the depth streams are synchronized for D435i cameras while their color cameras use the rolling shutter mechanism.
The arrangement of the three RGB-D cameras ensures a large perception FOV to form a ground-looking measurement system. 
Notably, the RGB-D cameras measure short-range, small-scale ground scenarios, 
while the LiDAR measures larger-scale surrounding scenarios.

\subsubsection{RTK}
An IMU-configured ZED-F9P RTK is employed, with its receiver installed at the rear of the sensor suite. 
It provides GNSS-IMU integrated trajectory readings\footnote{\url{https://github.com/jiminghe/Xsens_MTi_ROS_Driver_and_Ntrip_Client}} in the World Geodetic System 84 (WGS 84) reference frames.

\subsubsection{Robot Platforms}
Additionally, we capture the internal motion status of the robot, accounting for its specific mobility when navigating different locations with varying terrains. 
The messages types are custom definitions for the wheeled robot Scout mini \footnote{\url{https://github.com/agilexrobotics/scout_ros/tree/master/scout_msgs/msg}} and the quadruped robot Unitree Go1 \footnote{\url{https://github.com/unitreerobotics/unitree_ros_to_real/tree/master/unitree_legged_msgs/msg}}).
The robot profile encompasses built-in kinematics, odometry, and driving data, etc.
The wheeled robot employs high-speed CAN, while the legged robot utilizes Ethernet, linking to the NUC computer to ensure low-latency communication.

\begin{figure}[t!]
\vspace{0.3cm}
\centering
\includegraphics[scale=0.49]{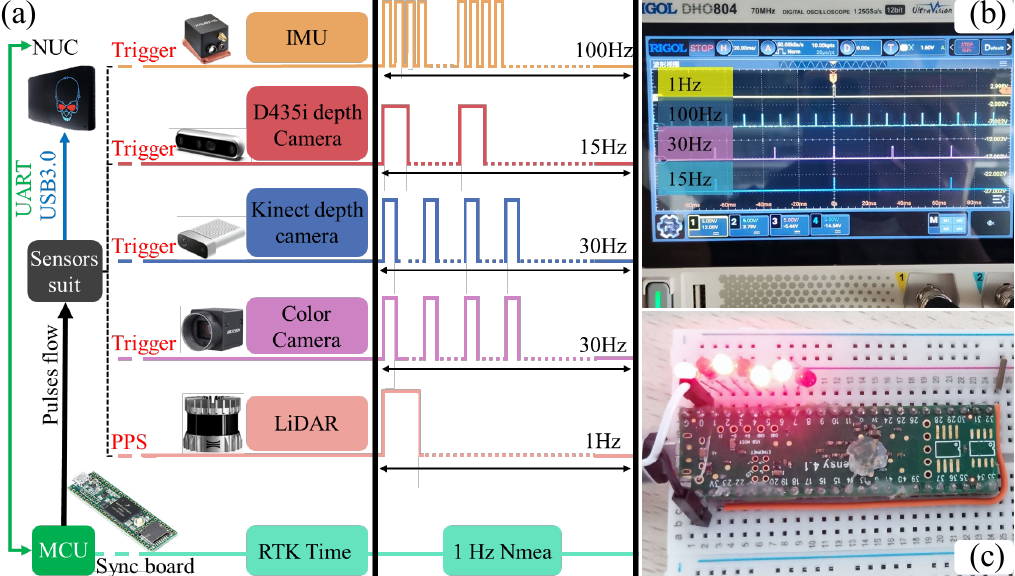}

\vspace{-0.2cm}
\caption{The description of sensor hardware time synchronization process. 
(a) Triggering implementation using different marked signals. (b) Visualization of the triggering pulses on the oscilloscope. (c) Validating the accuracy of hardware triggering using the Light Emitting Diode (LED) board.
}
\label{fig:sensor_sync}
\vspace{-0.4cm}
\end{figure}
\subsection{Time Synchronization}
\label{sec:synchronization}
We synchronize all sensors in the sensor suite at the hardware level to capture real-time data through the trigger signal in Fig.\ref {fig:sensor_sync}(a)(b).
We employ a Teensy 4.1 as the micro-controller unit (MCU) \cite{shi2022adaptive}.
It can obtain the National Marine Electronics Association (NMEA) pulse per second (PPS) signal from the RTK receiver to synchronize the ROS time from the GPS time.
To enhance the scalability of the hardware trigger system and mitigate response time variations among different cameras, the MCU generates 12 channels of trigger pulses at 3.3V.
Configuration of the frequency and time offset for these triggering pulses is adjustable through a serial debugging assistant on the host computer.
Furthermore, all trigger signals are accurately aligned to the PPS signal, exhibiting a precision of less than 1 microsecond.

Upon receiving a start/configuration pulse-per-second (PPS) signal from the RTK, the MCU dynamically programs various preconfigured trigger signals for the LiDAR, D435i cameras, Kinect, stereo frame cameras, and IMU.
The LiDAR's internal clock is synchronized through the PPS, initiating the capture of individual scans triggered by a 1Hz signal.
Following a 15Hz trigger pulse, only the depth camera in the RGB-D D435i is directly activated, while the color camera achieves data synchronization with the depth camera through its built-in software synchronization.
The stereo frame cameras are triggered by a 30Hz signal, activating the global shutter for the predetermined exposure duration.
The same 30Hz signal is also forwarded to the Kinect camera.
Additionally, the IMU is triggered by a sequence of 100 pulses at 10ms intervals via rising edges.
When receiving an end trigger signal from the command, all aforementioned signals, except PPS (from RTK), are terminated.

To assess the precision of sensor hardware synchronization, we conducted a flashing experiment depicted in Figure \ref{fig:sensor_sync}(c).
The MCU receives the PPS signal from RTK and generates seven channels of 1Hz pulses with a rising edge width of 0.45ms to drive LED flashes. 
To achieve a continuous change in the states of the LED captured by the cameras, the LED flashes sequentially in a 7-bit binary pattern, with each LED remaining illuminated for a duration of less than 0.45ms.
Multiple cameras can capture the LED light with the same binary state simultaneously, which indicates that the accuracy of our customized synchronization module can reach the millisecond level.

\begin{table}[t!]
\vspace{0.3cm}
\centering 
\caption{Specification of calibration process}

\vspace{-0.2cm}
\setlength{\tabcolsep}{2.5mm}{
\resizebox{1.0\columnwidth}{!}{
\begin{tabular}{ll}
\toprule[0.03cm]
Calibration Item & Methods \\ \bottomrule[0.03cm]
Intrinsic of IMU & \texttt{Allan Variance ROS} \\
Intrinsic of Color Cameras & \texttt{ROS camera calibration} \\ 
Stereo frame Cameras Extrinsic & \texttt{Stereo Camera} \\ 
Color Camera-IMU Extrinsic & \texttt{Kalibr} \\ 
Kinect Color Camera-LiDAR Extrinsic & \texttt{MATLAB} \\ 
\bottomrule[0.03cm]
\end{tabular}}}
\label{tab:sensor_calibration}
\vspace{-0.7cm}
\end{table}

\begin{figure*}[t!]
  \vspace{0.3cm}
  \centering
  \includegraphics[scale=0.52]{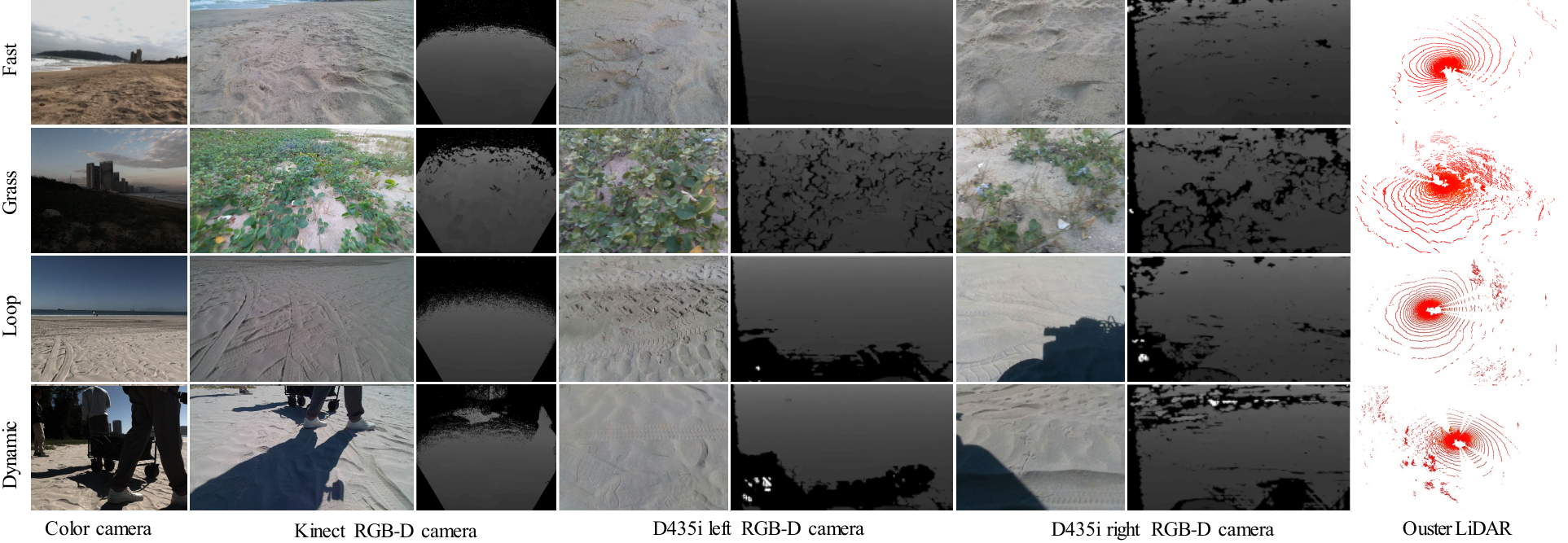}

  \vspace{-0.2cm}
  \caption{Sampled challenging scenarios shows the in-sequence diversity. We capture a variety of terrain characteristics with stereo frame cameras, three RGB-D cameras, and LiDAR. }
  \label{fig:sample_data}
  \vspace{-0.6cm}
\end{figure*}

\subsection{Sensor Calibration}
\label{sec:calib}
Table. \ref{tab:sensor_calibration} shows the illustration of all calibration variables and their calibration sequence.
Initially, the intrinsics parameters of each sensor are calibrated, followed by the estimation of extrinsics parameters with fixed intrinsics to establish spatial correlation.
We define the coordinate system of the Xsens IMU as the \textit{reference frame} where extrinsics of other sensors are referred to.

\subsubsection{Intrinsics calibration}
We calibrate the IMU intrinsic using the \texttt{Allan Variance ROS toolbox}\footnote{\url{https://github.com/ori-drs/allan_variance_ros}} and all the color cameras' intrinsic of pinhole models by the \texttt{ROS camera calibration toolbox}\footnote{\url{http://wiki.ros.org/camera_calibration}}.

\subsubsection{Extrinsics calibration}
We identify the spatial-temporal parameters between different color cameras with the Xsens IMU using the \texttt{Kalibr toolbox}\footnote{\url{https://github.com/ethz-asl/kalibr}}.
Note only the extrinsic between Xsens IMU concerning the color cameras are calibrated and defined as:
$\langle$ Xsens, frame cameras$\rangle$, 
$\langle$ Xsens, D435i color cameras$\rangle$, $\langle$ Xsens, Kinect color cameras$\rangle$.  
Additionally, we utilize the \texttt{MATLAB toolbox}\footnote{\url{https://www.mathworks.com/help/lidar/ug/lidar-and-camera-calibration.html}} to obtain the extrinsic for the Kinect color camera-LiDAR.
The \texttt{Stereo Camera toolbox}\footnote{\url{https://www.mathworks.com/help/vision/ug/using-the-stereo-camera-calibrator-app.html}} is used for the stereo frame cameras.

Other sensors can be derived from the calibration chain and transformations, or appended using the pre-existing, factory calibration parameters. 
Additionally, we incorporate the distinct coordinate systems of the robot platforms into the primary chain based on information provided in the manufacturer's manual. 
To aid future studies, we include Computer-Aided Design (CAD) measurements of all inter-sensor relative transformations for the actual assembly.

\section{Dataset Overview}
\label{sec:dataset}

This section introduces the detailed collection of diverse sequences, which are set as the criteria of the TAIL dataset.

\begin{figure}[t!]
  \vspace{0.2cm}
  \centering
  \includegraphics[scale=0.25]{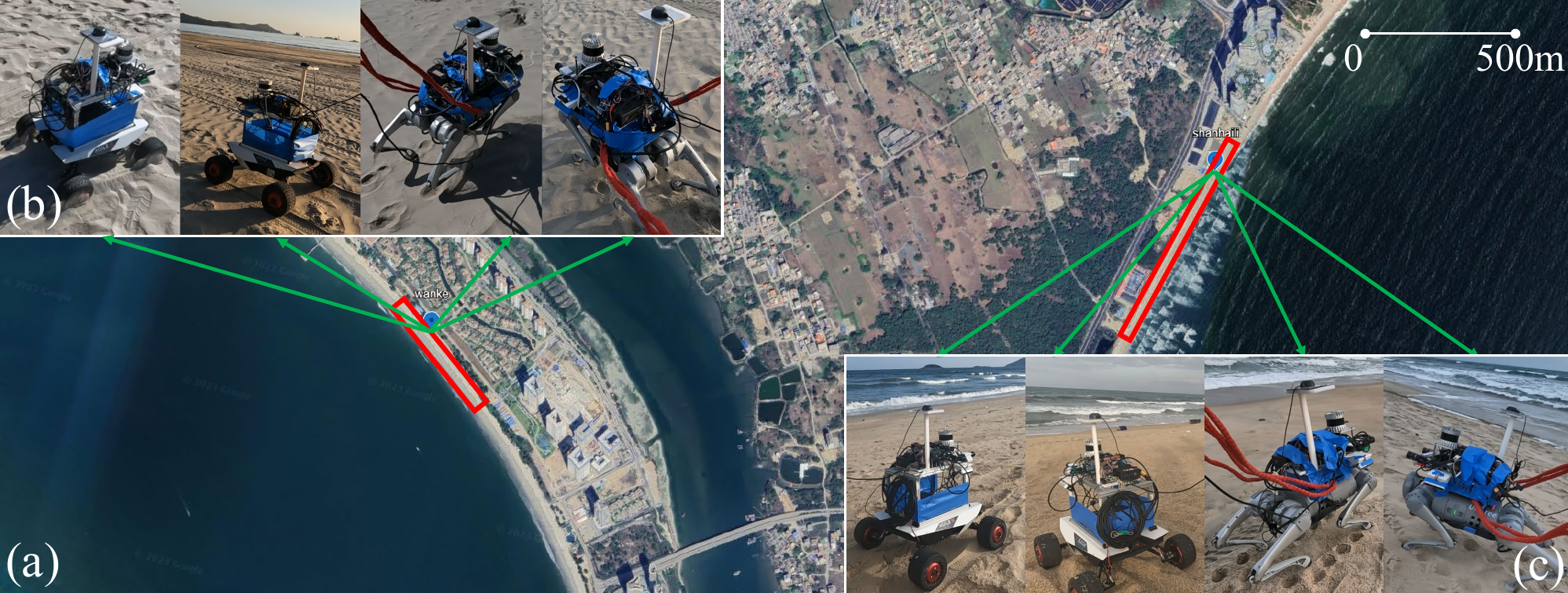}

  \vspace{-0.2cm}
  \caption{Overview of our field experiments for the TAIL data collection. (a) Acquisition locations are depicted on the satellite map. (b) Coarse, steep beach from \texttt{shanhaili}. (c) Fine, smooth beach from \texttt{wanke}.}
  \label{fig:scene_images}
  \vspace{-0.6cm}
\end{figure}

\begin{figure}[t!]
  \vspace{0.2cm}
  \centering
  \includegraphics[scale=0.4]{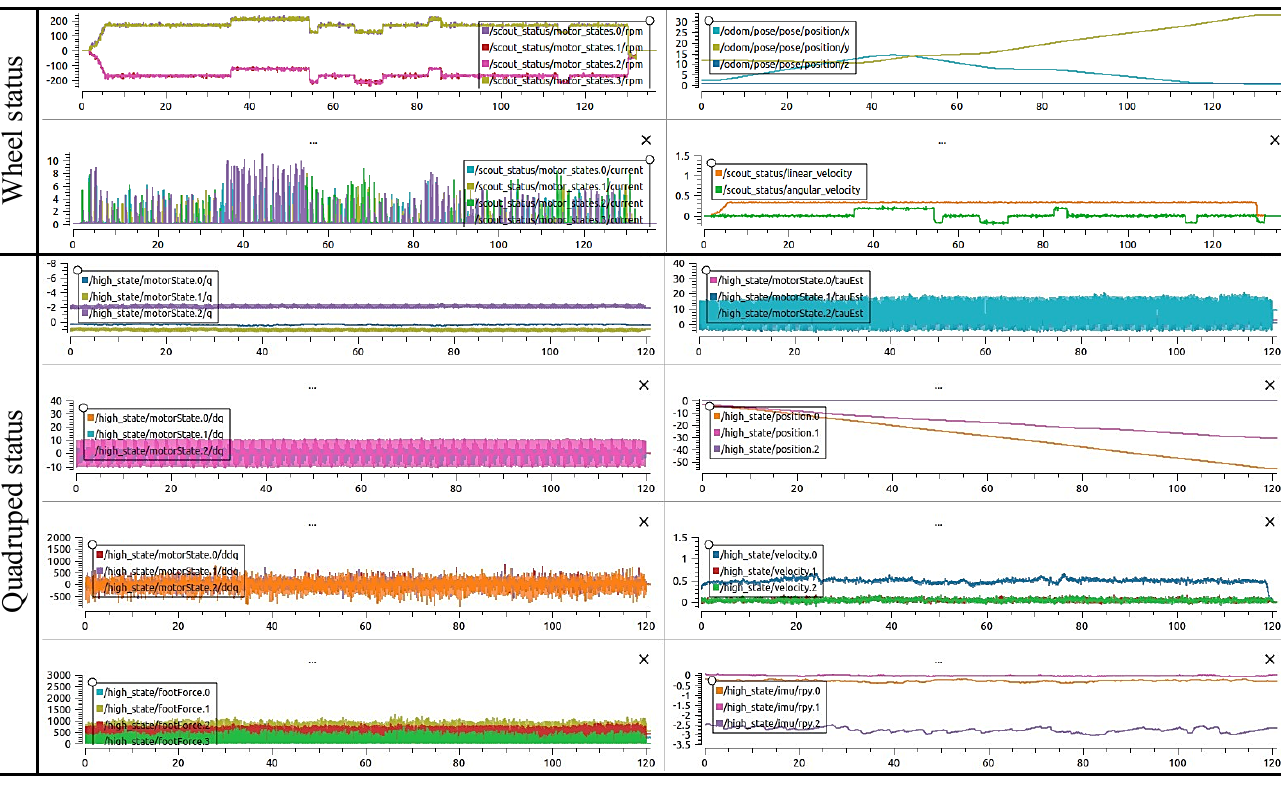}
  \caption{Visualization of different inertial kinematic parameters from wheeled and quadruped robots.}
  \label{fig:robot_stutus}
  \vspace{-0.6cm}
\end{figure}

\begin{table*}[t!]
\vspace{0.3cm}
\centering 
\caption{Overview of dataset sequences}

\vspace{-0.2cm}
\setlength{\tabcolsep}{2.0mm}{
\resizebox{2.0\columnwidth}{!}{
\begin{tabular}{llllcccccccc}
\toprule[0.03cm]
Platforms & Locations & Motions & Sequneces Description & Seq. & T[s] & $||\overline{\bm{v}}||[m/s]$ & $ \bm{L}$[m] & Lighting & Appearance & Objects & Size[GB] \\ \bottomrule[0.03cm]
\multirow{11}{*}
{\begin{tabular}[c]{@{}c@{}}
Wheeled \\ Robot
\end{tabular}} 
& \multirow{3}{*}{shanhaili} 
 & slow & backward, planar & 1 & 162 & 0.4 & 63.1 & weak & grassy, coarse sand & static & 21.8 \\
 &  & medium & forward, planar & 2 & 177 & 0.5 & 87.0 & normal & coarse sand & static & 21.3 \\ 
 &  & fast & forward, planar & 3 & 105 & 0.8 & 81.3 & normal & coarse sand & static & 12.3 \\ \cline{2-12} 
 & \multirow{6}{*}{wanke} & slow & forward, planar & 4 & 142 & 0.1 & 15.0 & weak & fine sand & static & 17.0 \\
 &  & slow & backaward, planar & 5 & 137 & 0.3 & 39.7 & normal & grassy, fine sand & static & 17.2 \\
 &  & slow & backaward, planar & 6 & 137 & 0.3 & 37.6 & normal & fine sand & dynamic &  15.1 \\ 
 &  & medium & backaward, planar & 7 & 145 & 0.4 & 58.5 & normal & fine sand & static &  16.3 \\
 &  & fast & forward, planar & 8 & 176 & 0.8 & 148.5 & strong & fine sand & static & 19.3 \\ \hline
\multirow{7}{*}
{\begin{tabular}[c]{@{}c@{}}
Quadruped \\ Robot 
\end{tabular}}  
& \multirow{2}{*}{shanhaili} & slow & forward, planar, jerky & 9 & 153 & 0.4 & 58.9 & weak & coarse sand & static & 17.4 \\
 &  & fast & forward, planar, jerky & 10 & 74 & 0.5 & 37.1 & weak & coarse sand & static &  8.5\\ \cline{2-12} 
 & \multirow{4}{*}{wanke} & slow & forward, planar, jerky & 11 & 144 & 0.3 & 49.6 & strong & fine sand & static &  15.1\\ 
 &  & fast & forward, planar, jerky & 12 & 148 & 0.7 & 103.3 & strong & fine sand & static &  15.1\\
 &  & fast & backward, planar, jerky & 13 & 175 & 0.3-0.7 & 69.4 & normal & fine sand & dynamic & 19.3 \\
 &  & fast & Multiple loops, planar, jerky & 14 & 194 & 0.5 & 90.5 & normal & fine sand & static &  21.3\\ 
\bottomrule[0.03cm]
\multicolumn{12}{l}{
Seq. means sequences.
$||\overline{\bm{v}}||$ means the commanded linear velocity.
$ \bm{L}$ means the total traveling distance.
T means total time.
}\\
\end{tabular}}}
\label{tab:dataset_summary}
\vspace{-0.6cm}
\end{table*}

\subsection{Dataset Scenarios}
\label{sec:dataset_sequence}
We provide 14 traveled sequences within the outdoor sites to test the performance of SLAM algorithms under various levels of challenges.
Two specific locations on the beaches of Double Moon Bay (22°N, 114°E), known as \texttt{shanhaili} and \texttt{wanke}, are selected as our field analog sites since they are natural, unstructured, and sandy with distinct size in widely-open areas.
Fig. \ref{fig:scene_images} reveals the filed scenarios for the diverse traverses, which have individual unique geological characteristics. 
Additionally, Fig. \ref{fig:sample_data} shows sensor measurements obtained from wheel and quadruped platforms in different scenarios.
Tab. \ref{tab:dataset_summary} concludes and describes the features of each sequence. 
Note both types of robot inertial kinematic status are collected, including dynamic motions under different types of terrain conditions. 
The characteristics are categorized as follows:
\begin{enumerate}
\item \textbf{Terrains:} Unstructured terrains can be found in coarse and fine sand benches, mainly ranging from easily traversable areas to sloped areas that are hard to cross. 
The granular mediums in such terrains have distinct particle sizes and specific mechanical properties, which influence SLAM with ground interaction \cite{yao2023wheel, xue2023wheel}.
\item \textbf{Surface appearance:} Granular sands tend to exhibit low saliency features and lack texture, which can influence visual analysis.
Moreover, the grassy, dry/wet sands offer a greater variety of ground features. 
\item \textbf{Illumination:} Color cameras are sensitive to lighting conditions, especially when the robot moves backlight or encounters issues such as underexposure or strong exposure. Such motions would negatively affect the visual processing algorithms.
\item \textbf{Dynamic objects:} Moving objects inevitably affect perception performance and leave "ghostly shadows" \cite{yao2021robust, zhu2022fusing}, which can not be treated as static background data for both visual and LiDAR perception. 
\item \textbf{Multiple loops:} Multiple loop trajectories are tested by driving in a circular path several times, which can help validate the accuracy of the odometry.
\item \textbf{Motion patterns:} Different motion patterns (low, medium, high speeds) can result in significant camera blur and terrain hazards \cite{yao2023predict, shi2022adaptive}. The corresponding kinematic parameters regarding different robot motions can be seen in Fig. \ref{fig:robot_stutus}.
Moreover, wheeled robots exhibit planar motions while quadruped robots perform planar but jerky motions on granular sand. 
Such robotic interaction provides a complement of heterogeneous proprioception information for exteroception perception.
\end{enumerate}

\begin{figure}[t!]
  \vspace{0.2cm}
  \centering
  \includegraphics[scale=0.3]{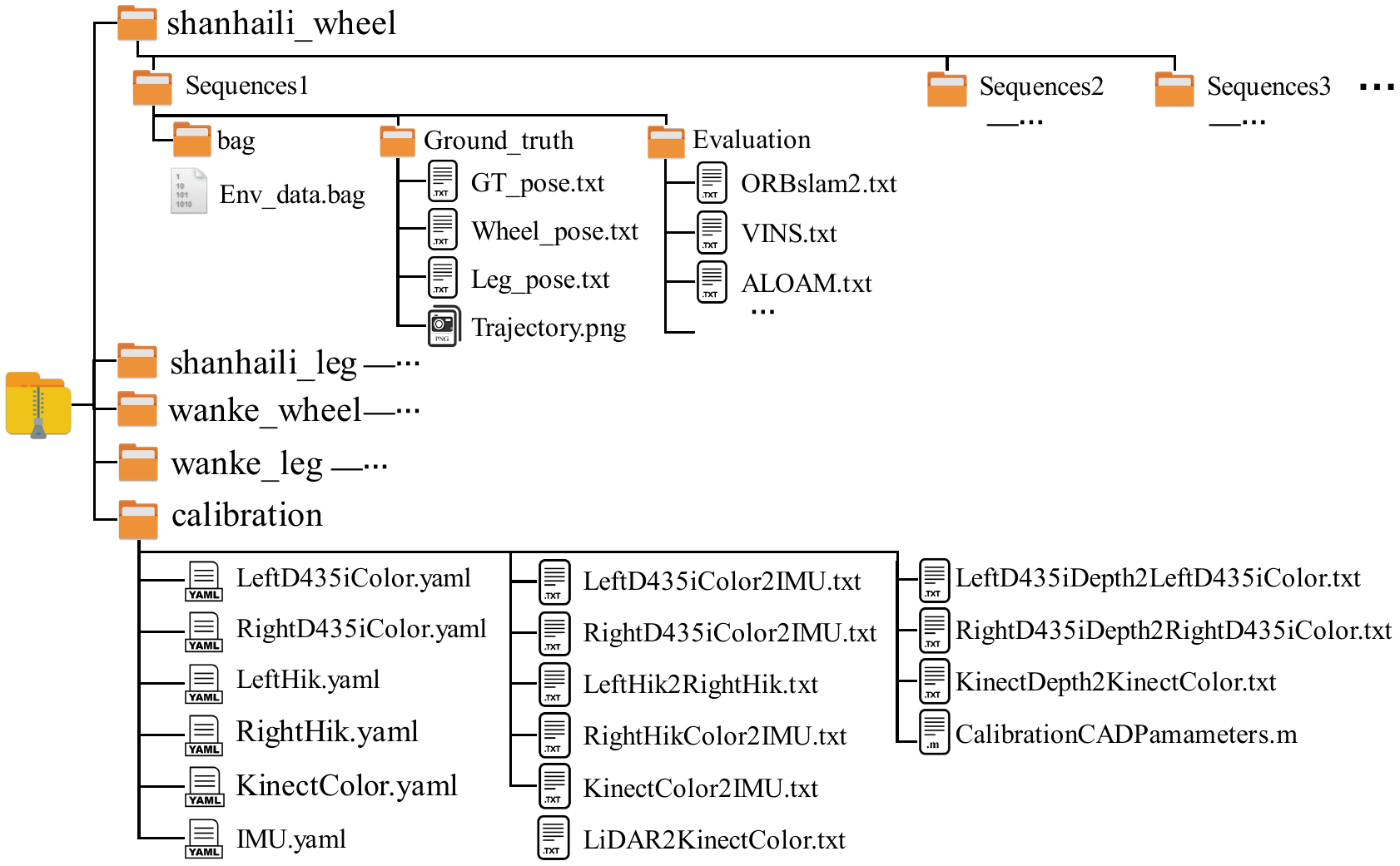}

  \vspace{-0.2cm}
  \caption{Structure of the TAIL dataset. In each package, the sequence is named as the observed environmental and the recorded date: \texttt{env\_date}. Ground-truth poses, odometry, and estimated trajectories are also contained.}
  \label{fig:dataset_format}
  \vspace{-0.6cm}
\end{figure}

\vspace{-0.3cm}
\subsection{Ground Truth}
\label{sec:dataset_groundtruth}
We provide 6-DoF (Degree of Freedom) reference ground truth for each sequence to evaluate the SLAM in soft terrains,  which is obtained by the IMU-integrated RTK measurements.
As SLAM methods represent poses concerning an arbitrary initial starting frame, we transform the pose trajectories obtained from diverse SLAM readings and IMU-integrated RTK into the LiDAR frame using extrinsic parameters. 
The aligned trajectories serve as a means of comparing the results with the ground truth for the latter evaluation in Sec. \ref{sec.experiment}.

\vspace{-0.3cm}
\subsection{Dataset Formats}
\label{sec:dataset_format}
Fig. \ref{fig:dataset_format} details the structure of TAIL dataset.
The simultaneously collected sensor data is stored in the format of a set of individual rosbags in corresponding sequences-\texttt{i}, which can be processed using ROS tools.
Timestamps for all messages are derived from their publication time on ROS and can be found in the \texttt{header.stamp} field.
Each sub-package is named after the observed sites and a postfix indicating the different motion configuration (\texttt{shanghaili\_wheel, shanhaili\_leg, wanke\_wheel, and wanke\_leg}). 
It also provides ground-truth trajectories with robotic pose for benchmarking and validation purposes. 
Moreover, the data evaluation files are shown to process the measurements with different SLAM algorithms.
Additionally, the dataset includes calibration files and frame relationships derived from CAD measurements.

\begin{table*}[t!]
\vspace{0.3cm}
\centering 
\caption{Performance comparison of SOTA SLAM algorithms on our dataset}

\vspace{-0.2cm}
\setlength{\tabcolsep}{2.0mm}{
\resizebox{2.0\columnwidth}{!}{
\begin{tabular}{ccccccccccccccccc}
\toprule[0.05cm]
\multirow{2}{*}{Platforms} & \multirow{2}{*}{Locations} & \multirow{2}{*}{Seq.} & \multicolumn{2}{c}{Odometry} & \multicolumn{2}{c}{ORB-SLAM2} & \multicolumn{2}{c}{VINS} & \multicolumn{2}{c}{VINS-Loop} & \multicolumn{2}{c}{A-LOAM} & \multicolumn{2}{c}{FAST-LIO2} & \multicolumn{2}{c}{FAST-LIVO} \\
 &  &  & RPE & ATE & RPE & ATE & RPE & ATE & RPE & ATE & RPE & ATE & RPE & ATE & RPE & ATE \\\bottomrule[0.05cm]
\multirow{8}{*}
{\begin{tabular}[c]{@{}c@{}}
Wheeled \\Robot
\end{tabular}} 
& \multirow{3}{*}{\begin{tabular}[c]{@{}c@{}}shanhaili,\\ coarse sand\end{tabular}} & 1 & 0.008 & 0.548 & 0.020 & 2.472 & 0.135 & 2.678 & 0.091 & 2.666 & 0.025 & 0.360 & 0.018 & \textbf{0.287} & 0.024 & 0.554 \\
 &  & 2 & 0.005 & 14.174 & 0.014 & 5.412 & 0.018 & 6.034 & 0.041 & 5.582 & 0.050 & 1.796 & 0.021 & \textbf{0.799} & 0.017 & 0.950 \\
 &  & 3 & 0.008 & 15.242 & $\times$ & $\times$ & $\times$ & $\times$ & $\times$ & $\times$ & 0.089 & 10.486 & 0.043 & \textbf{1.282} & 0.063 & 3.138 \\ \cline{2-17} 
 & \multirow{5}{*}{\begin{tabular}[c]{@{}c@{}}wanke,\\ fine sand\end{tabular}} & 4 &  0.001 & 1.056 & 0.008 & \textbf{0.135} & 0.008 & 3.774 & 0.200 & 3.538 & 0.011 & 0.307 & 0.008 & 0.333 & 0.008 & 0.323 \\
 &  & 5 & 0.003 & 2.068 & 0.010 & 0.923 & 0.006 & 1.013 & 0.021 & \textbf{0.887} & 0.017 & 0.911 & 0.010 & 0.929 & 0.015 & 0.896 \\
 &  & 6 & 0.002 & 5.015 & 0.009 & 0.889 & 0.084 & 7.754 & 0.089 & 8.076 & 0.017 & \textbf{0.814} & 0.011 & 0.879 & 0.017 & 0.838 \\
 &  & 7 & 0.002 & 1.887 & 0.017 & 3.291 & 0.061 & 4.207 & 0.225 & 3.891 & 0.017 & 1.406 & 0.011 & 1.441 & 0.011 & \textbf{1.336} \\
 &  & 8 & 0.005 & 27.774 & $\times$ & $\times$ & 0.032 & 11.556 & 0.091 & 8.578 & 0.029 & \textbf{3.008} & 0.014 & 3.215 & 0.024 & 3.021 \\  \bottomrule[0.03cm]
\multirow{7}{*}
{\begin{tabular}[c]{@{}c@{}}
Quadruped \\Robot 
\end{tabular}} 
& \multirow{2}{*}{\begin{tabular}[c]{@{}c@{}}shanhaili,\\ coarse sand\end{tabular}} & 9 & 0.010 & \textbf{1.623} & $\times$ & $\times$ & $\times$ & $\times$ & $\times$ & $\times$ & 0.102 & 2.250 & 0.089 & 5.982 & 0.074 & 2.942 \\
&  & 10 & 0.011 & \textbf{0.828} & $\times$ & $\times$ & $\times$ & $\times$ & $\times$ & $\times$ & 0.083 & 4.982 & $\times$ & $\times$ & 0.197 & 9.598 \\ \cline{2-17} 
& \multirow{4}{*}{\begin{tabular}[c]{@{}c@{}}wanke,\\ fine sand\end{tabular}} & 11 & 0.007 & 1.263 & $\times$ & $\times$ & $\times$ & $\times$ & $\times$ & $\times$ & 0.041 & \textbf{0.737} & 0.067 & 1.201 & 0.140 & 0.758 \\
&  & 12 & 0.014 & 3.076 & $\times$ & $\times$ & $\times$ & $\times$ & $\times$ & $\times$ & 0.047 & \textbf{2.263} & 0.022 & 2.557 & 0.032 & 2.334 \\
&  & 13 & 0.009 & 2.486 & $\times$ & $\times$ & $\times$ & $\times$ & $\times$ & $\times$ & 0.038 & \textbf{1.334} & 0.023 & 1.559 & 0.026 & 1.453 \\
&  & 14 & 0.009 & \textbf{0.597} & $\times$ & $\times$ & $\times$ & $\times$ & $\times$ & $\times$ & 0.117 & 6.873 & 0.150 & 27.942 & 0.166 & 31.553 \\ \bottomrule[0.05cm]	
\multicolumn{17}{l}{
 "Loop" means that the loop closure is considered.
"$\times$" means that the method fail in this sequence. 
}
\end{tabular}}}
\label{tab:exp_localization_accuracy}
\vspace{-0.4cm}
\end{table*}

\begin{figure*}[t!]
\centering
\subfigure[Sequence 1]
{\label{fig:traj_02}\centering\includegraphics[width=.246\linewidth]{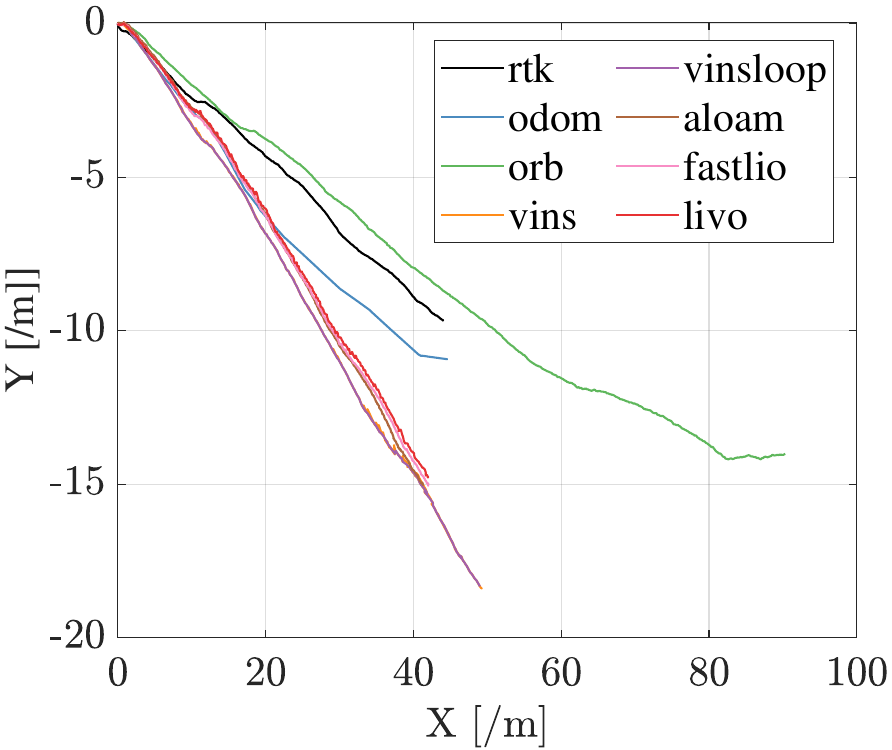}}
\subfigure[Sequence 8]
{\label{fig:traj_11}\centering\includegraphics[width=.24\linewidth]{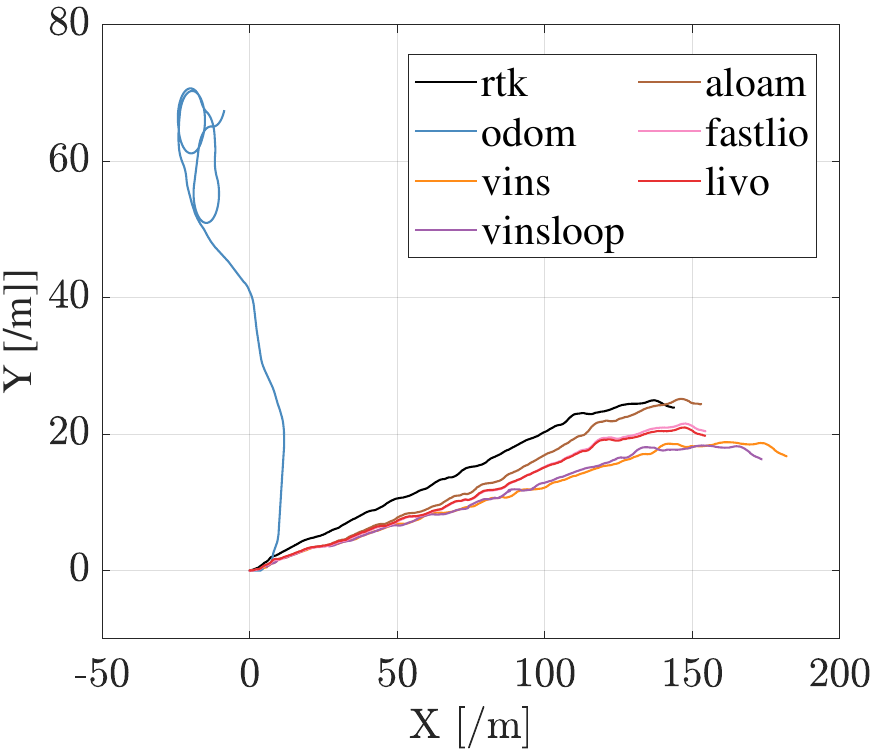}}
\subfigure[Sequence 9]
{\label{fig:traj_12}\centering\includegraphics[width=0.238\linewidth]{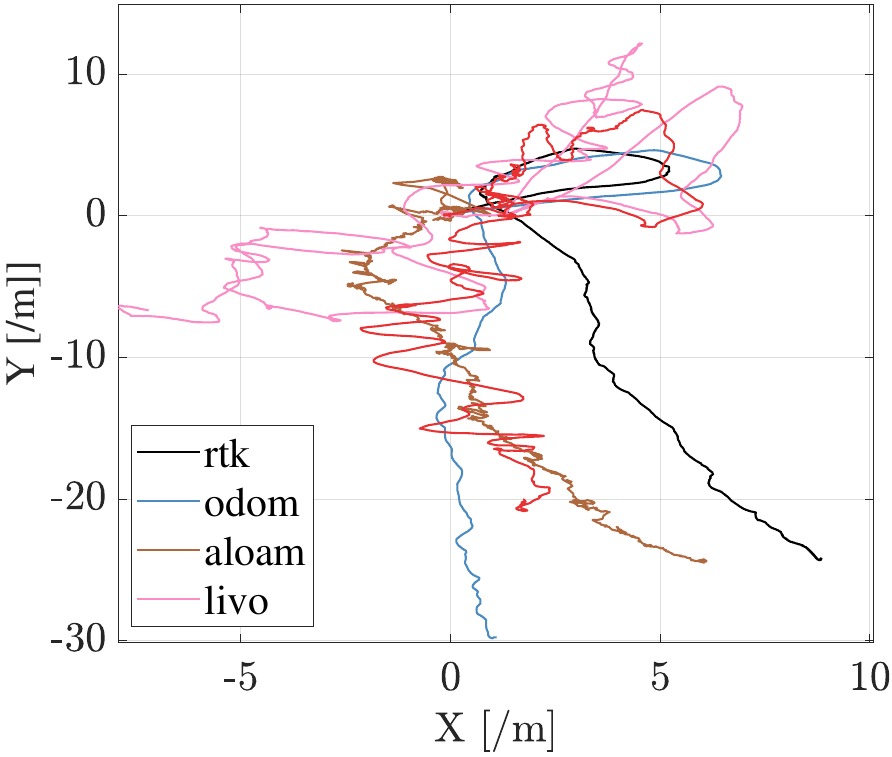}}
\subfigure[Sequence 14]
{\label{fig:traj_18}\centering\includegraphics[width=.24\linewidth]{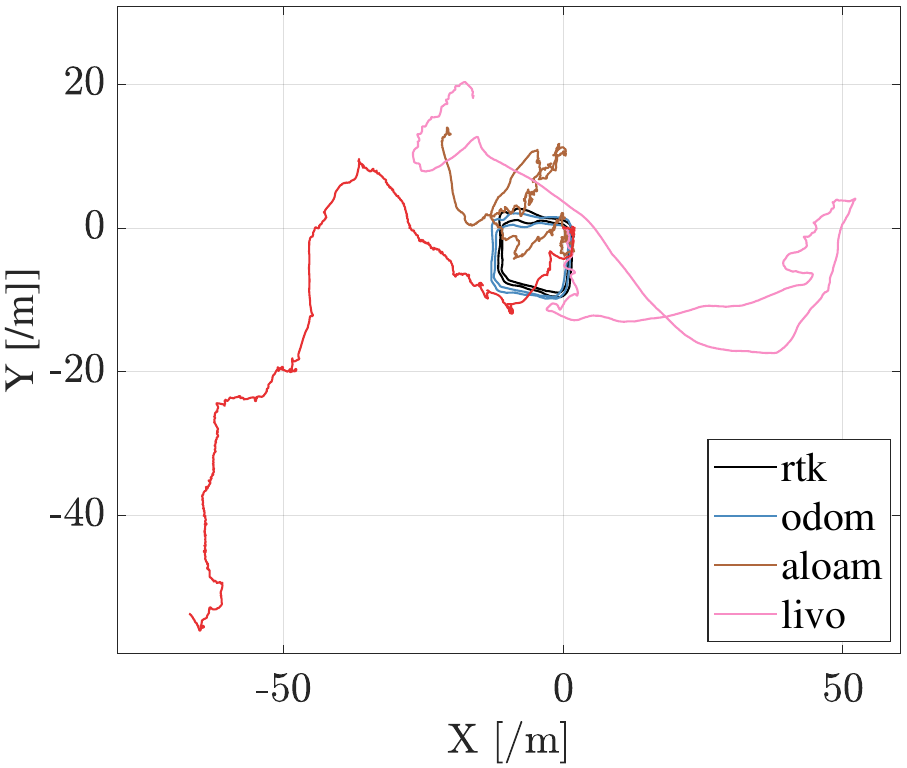}}

\vspace{-0.3cm}
\centering
\subfigure[Map visualization for the above corresponding sequences]
{\label{fig:mapvize}\centering\includegraphics[width=.98\linewidth]{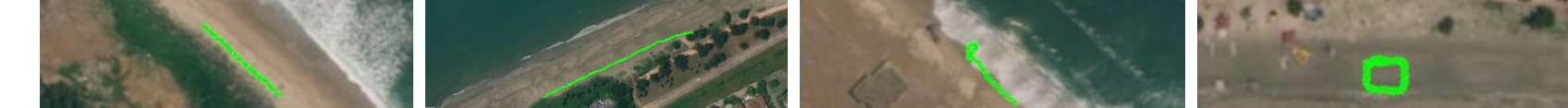}}

\vspace{-0.3cm}
\caption{Comparison of estimated trajectories with respect to the ground truth in four representative data sequences.
}
\label{fig:exp_traj}
\vspace{-0.5cm}
\end{figure*}  

\section{Experiment}
\label{sec.experiment}

To validate the versatility of the proposed datasets, we conduct a benchmarking study on several SOTA SLAM systems and report their results. 
The robots' internal odometry and the following open-source algorithms using diverse sensor combinations and technologies are evaluated: 
ORB-SLAM2 (right D435i color cameras) \cite{mur2017orb}, 
VINS-Fusion (IMU+left frame camera) \cite{qin2019general},
A-LOAM (LiDAR-only) \cite{zhang2014loam},
FAST-LIO2 (IMU+LiDAR) \cite{xu2022fast},
and FAST-LIVO (IMU+LiDAR+left frame camera) \cite{zheng2022fast}.
The default parameter configuration files are rectified to fit our sequences and also released.
We use \texttt{EVO}\footnote{\url{https://github.com/MichaelGrupp/evo}} with the Technical University of Munich (TUM) format (time, x,y,z,qx,qy,qz,qw) to analyze pose accuracy. 
The mean absolute trajectory error (ATE) and relative trajectory error (RTE) of estimation in the translation part are calculated w.r.t. the ground truth. 
Tab. \ref{tab:exp_localization_accuracy} summarizes the quantitative localization results, while Fig. \ref{fig:exp_traj} visualizes the estimated trajectories.

Our TAIL dataset can support a wide range of autonomy frameworks that vary significantly, including mono, visual-inertial, LiDAR-only, LiDAR-inertial, LiDAR-visual-inertial fusions, and robot odometry. These frameworks also demonstrate the quality of our synchronization and calibration.

The TAIL dataset can serve as a rigorous test bench and the ATE results are used as a baseline to elucidate inherent shortcomings.
LiDAR-based methods generally outperform vision-based methods when operating in fine terrains. Both methods, however, are prone to failure due to sensor degradation.	
ORB-SLAM2 and VINS-Fusion fail in some wheel robot driving cases and all quadruped robot traveling cases (8/14 tests fail). These failures occur when the scenes have dark or dynamic images while the scenes are texture-less, leading to poor initialization. Additionally, image blurring and the rolling shutter effect induced by the robot's aggressive movement in deformable terrains adversely impact visual detection.
A-LOAM performs better in sequences involving travel on fine sand, capitalizing on LiDAR's enhanced perception capabilities, but it struggles in coarse terrains.
Although FAST-LIO2 has a remarkable performance with the aid of the IMU, it occasionally produces unreliable results in certain sequences, particularly when used with quadruped robots as the unreliable sensors' measurements.
FAST-LIVO exhibits greater robustness but does not perform well in localization accuracy, even in slow sequences involving wheel robots where the multi-modal features are prone to failure in such texture-less terrains.
Conversely, leg odometers ensure reliable operation in select quadruped locomotion sequences. These sequences have high textural monotonicity and severe vibrations rendering visual-based methods ineffective, indicating limited robustness in multi-degraded environments.

The results indicate that the integration of multiple sensors is essential for achieving high accuracy and robustness in complex environments, particularly when considering the internal kinematic parameters.
In general, this dataset may not be a perfect collection \cite{nguyen2021ntu}, but it provides valuable real-world scenarios encountered by various types of robot platforms. These scenarios can serve as performance tests, feasibility references, and representative comparisons for evaluating sensor fusion mechanisms.
The TAIL dataset highlights the necessity of applying advanced SLAM techniques in the real-world for the research community. We hope that TAIL will contribute to the advancement of autonomous robots by providing timely and complementary scene data.

\section{Conclusion}
\label{sec.conclusion}

This paper presents TAIL, a novel terrain-aware multi-modal dataset targeting the development of SLAM algorithms capturing unique wheel and legged motion when traveling over two dynamic sandy scenes.
The comprehensive data collection in the wilds allows TAIL to examine the scenes on surrounding and supporting ground, track scene changes, analyze the motion proprioception of these platforms, and enhance understanding of robotic-centric terrain interactions.
We benchmark SOTA SLAM algorithms using our dataset with ground truth and identify additional challenges to be tackled for robot navigation over soft terrains.

In the future, we plan to periodically update and extend this dataset to include more comprehensive and challenging environments, covering more real-world scenarios that can contribute to the development of multi-sensor fusion SLAM within distinct robots. 
Additionally, we intend to explore more terrain-aware tasks, including terrain segmentation, ground interaction and motion estimation for different autonomous robots in different terrains.

\section*{ACKNOWLEDGMENT}
The authors express their sincere gratitude to the technicians from Ouster and Xsens for their valuable assistance in the project's progress. Moreover, we appreciate the open-source methods and the members of FusionPortable \cite{jiao2022fusionportable} and VECtor \cite{gao2022vector}, which facilitate the development of our proposed framework.

\bibliographystyle{IEEEtran}
\bibliography{ref}

\end{document}